\documentclass[10pt,twocolumn,letterpaper]{article}

\usepackage{cvpr}              
\usepackage[accsupp]{axessibility}

\usepackage{graphicx}
\usepackage{amsmath}
\usepackage{amssymb}
\usepackage{booktabs}
\usepackage{stfloats}
\usepackage{bm}

\usepackage[pagebackref,breaklinks,colorlinks,bookmarks=false]{hyperref}

\usepackage[capitalize]{cleveref}
\crefname{section}{Sec.}{Secs.}
\Crefname{section}{Section}{Sections}
\Crefname{table}{Table}{Tables}
\crefname{table}{Tab.}{Tabs.}

\usepackage{enumitem}
\setitemize[1]{itemsep=0pt,partopsep=0pt,parsep=\parskip,topsep=2pt}

\def\cvprPaperID{11632} 
\def\confName{CVPR}
\def\confYear{2022}

\begin{document}

\title{Thin-Plate Spline Motion Model for Image Animation}

\author{Jian Zhao\qquad  Hui Zhang \\School of Software, BNRist, Tsinghua University, Beijing, China\\
{\tt\small zhaojian20@mails.tsinghua.edu.cn}\quad {\tt\small huizhang@tsinghua.edu.cn}
}


\maketitle

\begin{abstract}
  Image animation brings life to the static object in the source image according to the driving video. Recent works attempt to perform motion transfer on arbitrary objects through unsupervised methods without using a priori knowledge. However, it remains a significant challenge for current unsupervised methods when there is a large pose gap between the objects in the source and driving images. In this paper, a new end-to-end unsupervised motion transfer framework is proposed to overcome such issues.
Firstly, we propose thin-plate spline motion estimation to produce a more flexible optical flow, which warps the feature maps of the source image to the feature domain of the driving image. Secondly, in order to restore the missing regions more realistically, we leverage multi-resolution occlusion masks to achieve more effective feature fusion. Finally, additional auxiliary loss functions are designed to ensure that there is a clear division of labor in the network modules, encouraging the network to generate high-quality images.
Our method\footnote{Our source code is publicly available: \href{https://github.com/yoyo-nb/Thin-Plate-Spline-Motion-Model}{https://github.com/yoyo-nb/Thin-Plate-Spline-Motion-Model}.} can animate a variety of objects, including talking faces, human bodies, and pixel animations. Experiments demonstrate that our method performs better on most benchmarks than the state of the art with visible improvements in motion-related metrics.
\end{abstract}

\begin{figure}[t]
  \centering

    \includegraphics[width=1\linewidth]{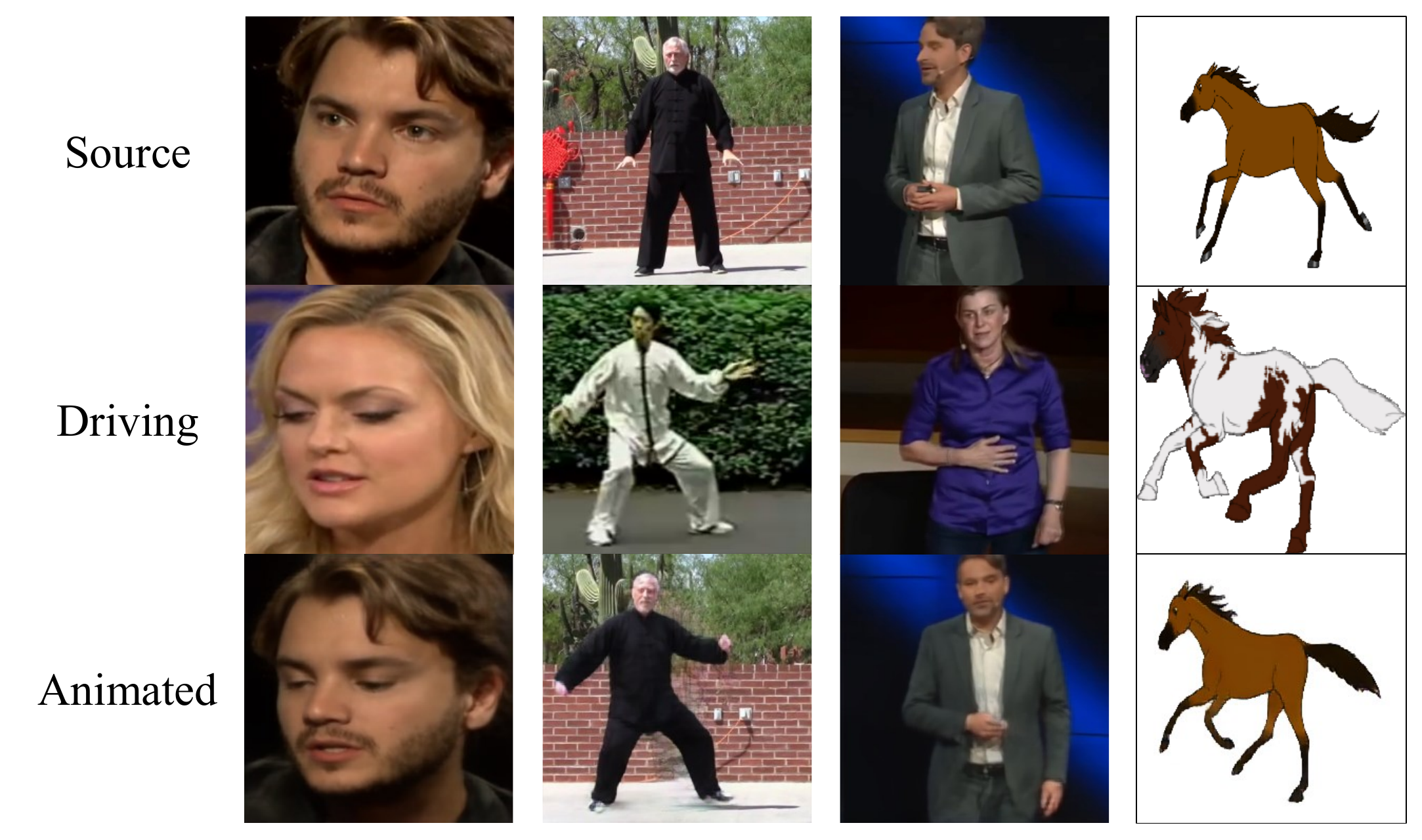}

    \caption{Example animations generated by our method trained on different datasets.}
    \label{fig:pic1}
    \vspace{-2mm}
\end{figure}

\section{Introduction}

Image animation (\cref{fig:pic1}) transfers the motion of the object in the driving video to the static object in the source image, which is widely used for video conferencing\cite{conferencing}, movie effects\cite{movie} and entertainment videos. It can stimulate people's creativity to create more interesting works.

Researches have been done on motion transfer by using a priori knowledge of objects, such as 3D models, landmarks, domain labels\cite{face2face,fewshot,marionette,headgan,everybody,3d_human,progressivepose,ren2020deep,skipgan}. However, these approaches, which rely on labeled data, only work for specific objects, such as faces\cite{face2face,fewshot,marionette,headgan} and human bodies\cite{everybody,3d_human,progressivepose,ren2020deep,skipgan}. It is costly to obtain such labeled data or pre-trained keypoint extractors. Therefore, these approaches cannot be applied to objects without labeled data.

Recently, some unsupervised motion transfer methods have been proposed that do not require a priori knowledge of objects\cite{x2face, monkey, fomm,mraa}. These methods use two frames sampled from a video for training, where one frame is used as the source image to reconstruct the other frame as the driving image. 
And the methods are optimized using reconstruction losses to learn the motion representations.  Some unsupervised methods\cite{monkey, fomm, mraa} divide motion transfer into two steps. First, an optical flow is estimated using the motion representation that warps the feature maps of the source image to the feature domain of the driving image. Second, an occlusion mask is predicted to indicate the missing regions of the warped feature maps, which are then inpainted in the network.
Experiments have shown that unsupervised methods can perform motion transfer on various objects\cite{monkey, fomm,mraa}.

However, there are still some challenges with the unsupervised methods. First, the motion representation is not flexible enough, making it difficult for the network to learn the large pose gap between the objects in the source and driving images during training.  This deficiency results in large discrepancies between the warped feature maps and the  feature domain of the driving image. Moreover, the area of the occlusion mask will increase, making motion transfer too dependent on the inpainting capability of the network, which leads to the second problem: inadequate inpainting capability of the network. Previous works\cite{monkey, fomm,mraa} did not take full advantage of features at different scales to inpaint the missing regions, so it is difficult to generate more realistic images.

Some unsupervised methods\cite{fomm,mraa} improve the quality of the animation by combining local affine transformations to estimate the motion. However, the affine transformation is linear, which makes it difficult to represent complex motions. In fact, the motions of objects are often not linear locally (for example, when people open their mouths, their lips are curved). To overcome this, we introduce a more flexible nonlinear transformation, thin-plate spline (TPS) transformation, to approximate the motion and propose a new end-to-end unsupervised motion transfer framework. First, we predict several sets of keypoints to generate TPS transformations and combine them with the affine background transformation\cite{mraa} to estimate the optical flow. Furthermore, we perform dropout for multiple TPS transformations during the early stage of training so that each TPS transformation contributes to the estimated optical flow. TPS motion estimation makes the estimated optical flow more flexible, stable and robust than previously estimated\cite{fomm, mraa}. Second, we predict occlusion masks for each layer of warped feature maps, making the feature maps have a different focus for more efficient feature fusion. Finally, we design the auxiliary loss functions to make each module have a clearer division of labor, encouraging the network to generate high-quality images. The proposed framework approximates the motion more accurately and has a stronger inpainting  capability. To summarize, the main contributions are as follows:
\begin{itemize}
  \item We present TPS motion estimation to approximate the motion from the source image to the driving image. In addition, we perform dropout on multiple TPS transformations before combining them during the early stage of training.

  \item We propose a new end-to-end unsupervised motion transfer framework. It warps the feature maps of the source image using the estimated optical flow and then leverages multi-resolution occlusion masks to indicate the missing regions for inpainting.  
  \item Experiments demonstrate that our method outperforms previous unsupervised motion transfer methods on various datasets, including talking faces, taichi videos, TED-talk videos and pixel animations. In particular, there is a visible improvement in motion-related metrics.
\end{itemize}


\section{Related Work}
\noindent
\textbf{Motion transfer}.   There are many supervised motion transfer methods that require a priori knowledge of moving objects, such as landmarks\cite{fewshot,marionette,everybody,progressivepose,ren2020deep,skipgan}, 3D models\cite{face2face, headgan,3d_human} or domain labels\cite{stargan}. Specially, GANimation\cite{ganimation} uses the Facial Action Coding System (FACS)\cite{facs} to describe facial expressions. However, these methods cannot be applied to new objects without labeled data, such as pixel animations.

As a comparison, unsupervised methods do not need to introduce a priori knowledge of the animated object during training\cite{x2face, monkey, fomm, mraa}. X2Face\cite{x2face} learns the identity representation of the source image by the embedding network, and then generates an optical flow to warp the embedded image. Some unsupervised methods attempt to model the motion representation and disentangle identity and pose from the image. Monkey-Net\cite{monkey} estimates optical flow for animating by predicting several pairs of unsupervised keypoints. Based on this, first order motion model (FOMM)\cite{fomm} performs first-order Taylor expansions near each keypoint and approximates the motion in the neighborhood of each keypoint using local affine transformations, which significantly improves the quality of motion transfer. Siarohin \etal proposed motion representations for articulated animation (MRAA)\cite{mraa}, which improves the shortcomings of FOMM\cite{fomm} and achieves state-of-the-art performance of unsupervised methods. MRAA\cite{mraa} uses PCA-based motion estimation, which has better quality in representing articulated motions (e.g., human body). In addition, it adds background motion estimation to eliminate the negative effects of camera motion. These unsupervised methods can perform motion transfer on arbitrary objects. Comparing these methods, our approach uses TPS motion estimation, which estimates optical flow more flexibly and works better for large-scale motions.

\noindent
\textbf{Multi-scale feature fusion}. Multi-scale feature fusion has proven to be effective in several tasks in computer vision, including keypoint prediction\cite{mtcnn,facealignment,hrnet}, segmentation\cite{unet,fcn,pspnet} and image generation\cite{msggan, lapgan,lapsrn}. Extensive experiments have demonstrated that different layers of the network focus on different levels of features\cite{lapgan,laptrans,lapsrn}. Lower scale feature maps focus on the overall patterns of the image, while larger scale feature maps emphasize detailed textures. The unsupervised motion transfer methods\cite{fomm, mraa} estimate an occlusion mask for the missing regions of the warped feature maps and inpaint them through feature fusion ways. FOMM\cite{fomm} uses an hourglass network to upsample the warped feature maps gradually to reconstruct the driving image. While Monkey-Net\cite{monkey} and MRAA\cite{mraa}  warp multi-scale feature maps and add them to the decoder part of the hourglass network via skip connections\cite{skipgan}. However, they use a single occlusion mask for feature maps at different scales, which is not conducive to the network that learns features with different focuses at multiple scales. Our approach also uses a skip-connected hourglass network for inpainting. The difference is that we estimate multi-resolution occlusion masks for feature maps at different scales, allowing features to be more fully fused for more realistic inpainting.

\begin{figure*}
\centering
\includegraphics[scale=0.5]{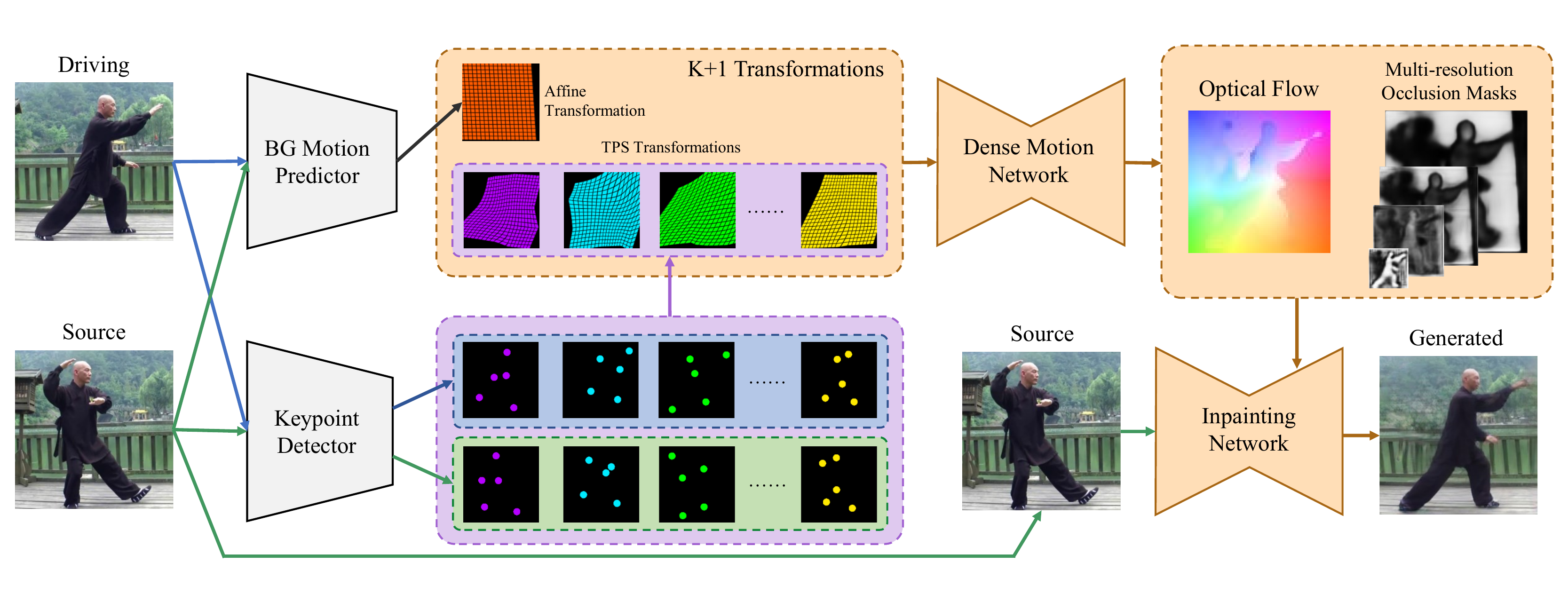}
\vspace{-3mm}
\caption{Overview of our model. The BG Motion Predictor predicts the affine transformation, representing the background motion from the source image to the driving image. At the same time, we estimate $K$ sets of keypoints using the Keypoint Detector, each of which generates a TPS transformation. The Dense Motion Network will then combine the $K+1$ transformations ($K$ TPS transformations and one affine transformation) for estimating optical flow and multi-resolution occlusion masks. Finally, we feed the source image into the Inpainting Network, warp the feature maps extracted by the encoder using optical flow, and mask them with the corresponding resolution occlusion masks. The generated image will be output at the last layer of the Inpainting Network.}
\label{fig:main}

\end{figure*}

\section{Method}
\cref{fig:main} shows the overview of our proposed model. It generates the reconstructed driving image $\hat{\mathbf{D}}$ given a source image $\mathbf{S}$ and a driving image $\mathbf{D}$. The model consists of following modules:

\begin{itemize}
 \item \textbf{Keypoint Detector}. The Keypoint Detector $E_{kp}$ receives $\mathbf{S}$ and $\mathbf{D}$ to predict $K\times N$ pairs of keypoints to generate $K$ TPS transformations.

 \item \textbf{BG Motion Predictor}. The concatenation of $\mathbf{S}$ and $\mathbf{D}$ is fed into the BG Motion Predictor $E_{bg}$ to estimate the parameters of the affine background  transformation.

 \item \textbf{Dense Motion Network}. The module is an hourglass network. It receives $K$ TPS transformations from $E_{kp}$ and one affine transformation from $E_{bg}$. The optical flow will be estimated by combining these $K+1$ transformations. At the same time, the multi-resolution occlusion masks are predicted by different layers of the decoder part to indicate missing regions of the warped feature maps.

 \item \textbf{Inpainting Network}. It is also an hourglass network. It warps the feature maps of the source image using the estimated optical flow and inpaints the missing regions of the feature maps for each scale. The generated image is output at the last layer.
\end{itemize}

\subsection{TPS Motion Estimation}
Motion estimation aims to approximate the mapping $\mathcal{T}$ such that $\mathcal{T}(\mathbf{S}) = \mathbf{D}$. In contrast to the combination of local affine transformations in FOMM\cite{fomm} and MRAA\cite{mraa}, we propose TPS motion estimation to approximate $\mathcal{T}$. 

TPS transformation\cite{tps} is a flexible, nonlinear transformation that allows representing more complex motions. Given corresponding keypoints in two images, we can warp one to the other with minimum distortion by using TPS transformation $\mathcal{T}_{tps}$:
\begin{equation}
  \label{eq:1}
  \begin{matrix}
    \begin{aligned}
      \min \iint_{R^{2}}\left(\left(\frac{\partial^{2} \mathcal{T}_{tps}}{\partial x^{2}}\right)^{2}\right.+2\left(\frac{\partial^{2} \mathcal{T}_{tps}}{\partial x \partial y}\right)^{2} \\ +\left. \left(\frac{\partial^{2} \mathcal{T}_{tps}}{\partial y^{2}}\right)^{2}\right) dx dy,
    \end{aligned}\\ 
    \text{s.t.} \quad \mathcal{T}_{tps}(P^{\mathbf{S}}_i) = P^{\mathbf{D}}_i, \quad  i=1,2, \ldots, N,
  \end{matrix}
\end{equation}
where $P^{\mathbf{X}}_i$ is the keypoints of image $\mathbf{X}$. We use the Keypoint Detector to predict $K\times N$ keypoints for $\mathbf{S}$ and $\mathbf{D}$, where $K$ is the number of TPS transformations. Every $N$ pairs ($N=5$ for this paper) of keypoints generate one TPS transformation from $\mathbf{S}$ to $\mathbf{D}$. According to the derivation in\cite{tps} , the $k^{th}$ TPS transformation is obtained as follows:
\begin{equation}
  \mathcal{T}_k(p)=A_k\begin{bmatrix}
    p\\ 
    1   
    \end{bmatrix}+\sum_{i=1}^{N} w_{ki} U\left({\left\lVert P^{\mathbf{D}}_{ki}-p\right\rVert}_2\right),
\end{equation}
where $p=(x, y)^\top$ is pixel coordinates, $A_k \in \mathcal{R}^{2\times 3}$ and $w_{ki} \in \mathcal{R}^{2\times 1}$ are the TPS coefficients obtained by solving \cref{eq:1}, $U(r)$ is a radial basis function, which represents the influence of each keypoint on the pixel at $p$:
\begin{equation}
  U(r)=r^{2} \log r^{2}.
\end{equation}

Besides, the camera motion in videos will cause the predicted keypoints to appear in the background area, leading to deviations in the motion estimation. To address this problem, we additionally predict an affine background transformation like MRAA\cite{mraa} to model the background motion:
\begin{equation}
  \mathcal{T}_{bg}(p) = A_{bg}\begin{bmatrix}
  p\\ 
  1
  \end{bmatrix},
\end{equation}
where $A_{bg}\in \mathcal{R}^{2\times 3}$ is an affine transformation matrix predicted by the BG Motion Predictor. 

Now, we will combine the $K+1$ transformations ($K$ TPS transformations and one affine transformation) to approximate the mapping $\mathcal{T}$. We use the $K+1$ transformations to warp $\mathbf{S}$, cascade the warped images and feed them into the Dense Motion Network. The module predicts $K+1$ contribution maps $\widetilde{\mathbf{M}}_k \in \mathcal{R}^{H\times W}, k=0,\dots,K$, where $H$ and $W$ are the height and width of the image, $\mathbf{\widetilde{M}}_0$ corresponds to $\mathcal{T}_{bg}$. The contribution maps are activated by softmax to make them sum to $1$ at any pixel location: 
\begin{equation}
  \mathbf{M}_k(p) = \frac{\exp(\widetilde{\mathbf{M}}_k(p))}{\sum_{i=0}^K\exp(\widetilde{\mathbf{M}}_i(p))}, k=0,\dots,K,
  \label{eq:no_drop}
\end{equation}
where $\mathbf{M}_k(p)$ is the value of $\mathbf{M}_k$ at coordinate $p$. We use $\mathbf{M}_k, k=0,\dots,K$ to combine the $K+1$ transformations to compute the optical flow:
\begin{equation}
  \widetilde{\mathcal{T}}(p) = \mathbf{M}_0(p)\mathcal{T}_{bg}(p)+ \sum_{k=1}^{K}\mathbf{M}_k(p)\mathcal{T}_k(p),
  \label{eq:warp}
\end{equation}
which is the result of our approximate mapping $\mathcal{T}$. We use $\widetilde{\mathcal{T}}$ to warp the feature maps of $\mathbf{S}$ extracted by the encoder of the Inpainting Network and reconstruct $\mathbf{D}$ in the decoder.

\noindent
\textbf{Dropout for TPS transformations}.
We use $K$ TPS transformations to approximate the motion, but only a few of them may work for the estimated optical flow at the early stage of training. Their contribution maps have zero values at any pixel location after softmax, and will have no contribution during the entire training stage. Therefore, the network can easily fall into local optimums, resulting in poor quality of the generated images.

We use dropout\cite{dropout} for TPS transformations to avoid this, which is a technique for regularization. Specifically, $\exp(\widetilde{\mathbf{M}}_k(p)), k=1,\dots,K$ are set to zero respectively with probability $P$ in softmax, such that some of $K$ TPS transformations do not work for estimating optical flow in this mini-batch training. And the terms not set to zero are divided by $1-P$ to ensure that the expectation of $\sum_{i=1}^K \exp(\widetilde{\mathbf{M}}_i(p))$ remains constant. Let $b_i, i=1,\dots, K$ obeys Bernoulli distribution with parameter $1-P$, we change \cref{eq:no_drop} by:

\begin{equation}
  \mathbf{M}_k(p) = \left\{\begin{matrix}
     \exp(\widetilde{\mathbf{M}}_0(p))/\mathbf{M}_T(p),  &k=0
    \\ 
    b_k \exp(\widetilde{\mathbf{M}}_k(p))/(1-P) \mathbf{M}_T(p) ,  &else
    \end{matrix}\right. ,
\end{equation}
where
\begin{equation}
  \mathbf{M}_T(p) = \exp(\widetilde{\mathbf{M}}_0(p))+\sum_{i=1}^K b_i \exp(\widetilde{\mathbf{M}}_i(p))/(1-P).
\end{equation}
Dropout keeps the network from excessive reliance on a few TPS transformations in the early stage of training, and increases the robustness of the network. 
We remove the dropout operation when each TPS transformation contributes to the estimated optical flow after training several epochs.

\begin{figure}
  \centering
  \includegraphics[scale=0.55]{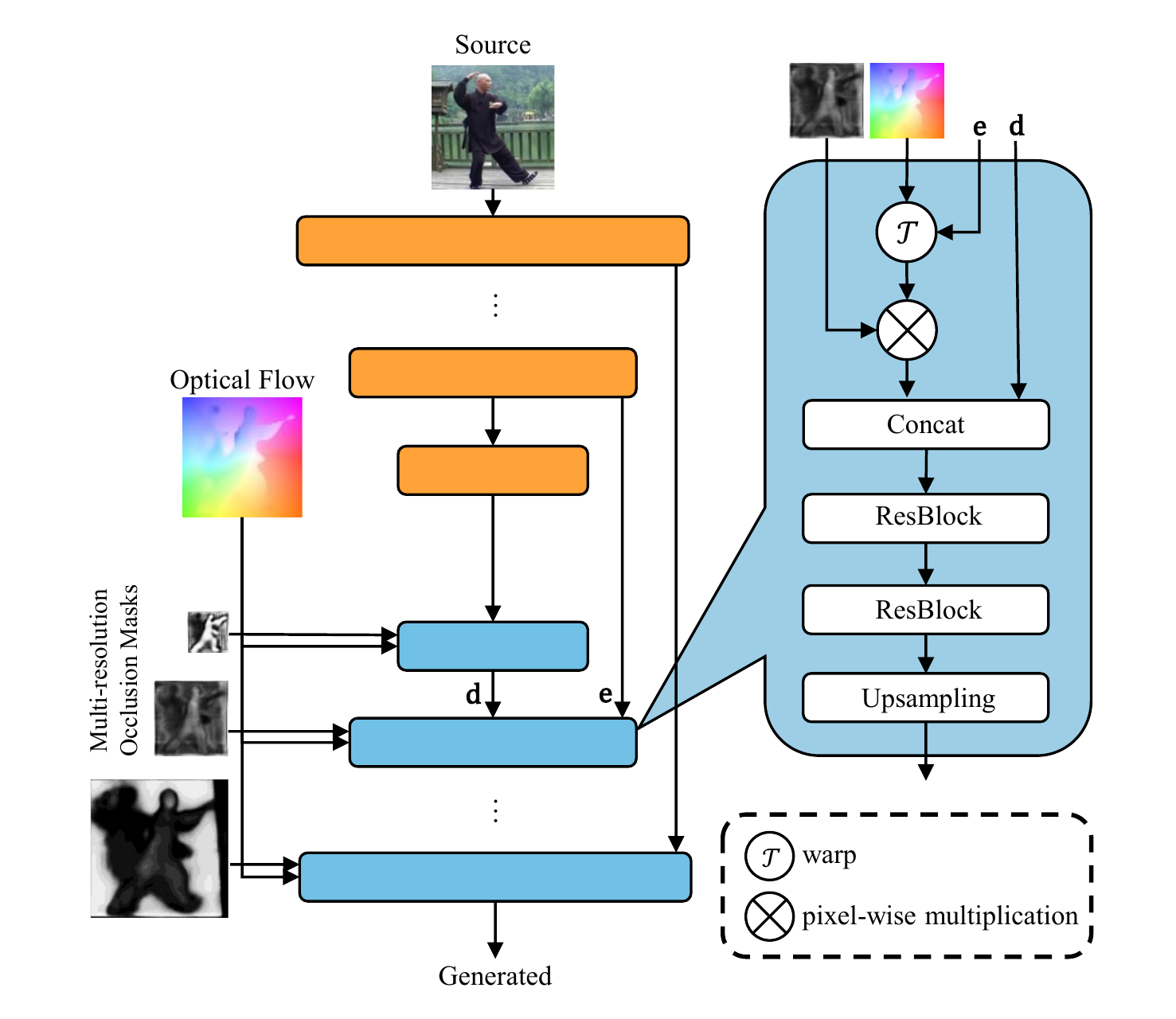}
  \caption{Implementation details of the Inpainting Network. }
  \label{fig:inpainting}
  \vspace{-2.mm}
\end{figure}

\subsection{Multi-resolution Occlusion Masks}

For the Dense Motion Network and the Inpainting Network, we employ the hourglass architecture network to fuse features of different scales, which has been proven effective in various works\cite{facealignment, laptrans,unet}. In \cite{fomm,mraa}, Dense Motion Network estimates a single occlusion mask for the warped feature maps to inpaint the missing regions. However, many experiments have shown that the focus of the feature map changes with its scale\cite{lapgan,laptrans,lapsrn}. Low-scale feature maps focus on abstract patterns, while high-scale feature maps are more concerned with detailed textures. If a single occlusion mask is used to mask out the feature maps of different scales, what it learns during training will be the trade-off among the focus of different scale feature maps. Hence, we predict occlusion masks of different resolutions for each layer of the feature map. The Dense Motion Network will not only estimate the optical flow but also predict the multi-resolution occlusion masks by using an additional convolution layer at each layer of the decoder. The estimated optical flow and the multi-resolution occlusion masks are fed together into the Inpainting Network.

In the Inpainting Network, we fuse multi-scale features to generate high-quality images, the details are shown in \cref{fig:inpainting}. We feed $\mathbf{S}$ into the encoder and use optical flow $\widetilde{\mathcal{T}}$ to warp the feature map of each layer. The warped feature map is then masked using the occlusion mask of corresponding resolution and is concatenated to the decoder part via a skip connection. The feature map is then upsampled after passing through two residual blocks. The generated image is output at the final layer.

\subsection{Training Losses}
Following FOMM\cite{fomm} and MRAA\cite{mraa}, we use a pre-trained VGG-19\cite{perceptual} network to calculate the reconstruction loss between $\mathbf{D}$ and the generated image $\hat{\mathbf{D}}$ at multi-resolutions:
\begin{equation}
\mathcal{L}_{rec}=\sum_{j}\sum_{i}\left|V_{i}(\mathbf{D}_j)-V_{i}(\hat{\mathbf{D}}_j)\right|,
\end{equation}
where $V_{i}$ is the $i^{th}$ layer of the pre-trained VGG-19 network, and $j$ represents that the image is downsampled j times. Equivariance loss is also used to constrain the Keypoint Detector in\cite{fomm,mraa}:
\begin{equation}
\mathcal{L}_{eq}=\left|E_{kp}(\mathcal{T}_{ran}(\mathbf{S}))-\mathcal{T}_{ran}(E_{kp}(\mathbf{S}))\right|,
\end{equation}
where $\mathcal{T}_{ran}$ is the random nonlinear transformation. We use the random TPS transformation like FOMM\cite{fomm} and MRAA\cite{mraa}. In addition, we designed auxiliary loss functions for the modules, namely bg loss and warp loss.

We used an affine transformation to model the background motion, and we made additional constraints on the BG Motion Predictor to make the predicted parameters more accurate and stable. We cascade in the order of $\mathbf{S}$ and $\mathbf{D}$ and feed them into the BG Motion Predictor to obtain the affine transformation matrix $A_{bg}$, representing the background's motion from $\mathbf{S}$ to $\mathbf{D}$. We then re-cascade them in reverse order to obtain the affine transformation matrix $A'_{bg}$. We expect the two affine transformation matrices to remain consistent:
\begin{equation}
  \label{eq:bg}
  \left[\begin{matrix}
     A'_{bg}   \\
    0\ \ 0\ \ 1
    \end{matrix}\right] = 
    \left[\begin{matrix}
      A_{bg}   \\
      0\ \ 0\ \ 1
     \end{matrix}\right]^{-1}.
\end{equation}
However, we cannot use \cref{eq:bg} as a loss function because it is easy to make the BG Motion Predictor output a zero matrix that minimizes the difference between the two sides of the equation. We reformulate \cref{eq:bg} in the following way:
\begin{equation}
  \vspace{-1.5mm}
  \mathcal{L}_{bg} = \left|\left[\begin{matrix}
    A'_{bg}   \\
   0\ \ 0\ \ 1
   \end{matrix}\right]\left[\begin{matrix}
    A_{bg}   \\
    0\ \ 0\ \ 1
   \end{matrix}\right]- I \right|,
\end{equation}
where $I$ is $3 \times 3$ identity matrix.

An additional constraint is also designed for the Inpainting Network. We feed $\mathbf{D}$ into the encoder of the Inpainting Network. The warped encoder feature maps of $\mathbf{S}$ are used to compute the loss with the encoder feature maps of $\mathbf{D}$ at each layer:

\vspace{-3mm}
\begin{equation}
  \mathcal{L}_{warp}=\sum_{i}\left|\widetilde{\mathcal{T}}(E_i(\mathbf{S}))-E_i(\mathbf{D})\right|,
\end{equation}
where $E_i$ is the $i^{th}$ layer of the Inpainting Network encoder. $\mathcal{L}_{warp}$ can encourage the network to estimate the optical flow more reasonably, making the warped feature maps closer to the feature domain of $\mathbf{D}$. 

The final loss is the sum of terms:
\vspace{-1.5mm}
\begin{equation}
\mathcal{L} = \mathcal{L}_{rec}+\mathcal{L}_{eq}+\mathcal{L}_{bg}+\mathcal{L}_{warp}.
\end{equation}

\subsection{Testing Stage}
At the testing stage, we use a source image $\mathbf{S}$ and a driving video $\{\mathbf{D}_t\}, t = 1,2,\dots ,T$ for image animation. FOMM\cite{fomm} has two modes for image animation: \emph{standard} and \emph{relative}. The \emph{standard} mode uses each frame $\mathbf{D}_t$ and $\mathbf{S}$ directly to estimate the motion using \cref{eq:warp}, while the \emph{relative} mode estimates the motion between $\mathbf{D}_t$ and the first frame $\mathbf{D}_1$ and then applies it to $\mathbf{S}$. However, the \emph{standard} mode does not perform well when there is a large mismatch between identities (for example, animate a thin person according to the motion of a fat person). The \emph{relative} mode requires that the pose of $\mathbf{D}_1$ be similar to the pose of $\mathbf{S}$. MRAA\cite{mraa} proposes a new mode, animation via disentanglement (\emph{avd}), that uses an additional trained network to predict the motion that should be applied to $\mathbf{S}$, and we use the same mode for image animation.

We train a shape and a pose encoder as in MRAA\cite{mraa}. The shape encoder learns the shape of the keypoints of $\mathbf{S}$, and the pose encoder learns the pose of the keypoints of $\mathbf{D}_t$. Then a decoder reconstructs the keypoints, preserving the shape of $\mathbf{S}$ and the pose of $\mathbf{D}_t$. Both the encoders and the decoder are implemented by fully connected layers. We use keypoints of two frames from a video to train the networks, where the keypoints of one frame are randomly deformed to simulate the pose of a different identity. $\mathcal{L}_1$ loss is used to encourage the networks to reconstruct the keypoints before deformation. For image animation, we feed the keypoints of $\mathbf{S}$ and $\mathbf{D}_t$ into the shape and pose encoders to get the reconstructed keypoints, and then use \cref{eq:warp} to estimate the motion.

\begin{table*}[!ht]
  \centering
  \resizebox{\textwidth}{!}{
  \begin{tabular}{c|ccc|ccc|ccc|c}
    
   \toprule
               &       & TaiChiHD       &       &       & TED-talks     &       &       & VoxCeleb &       & MGif   \\
               & $\mathcal{L}_1$    & (AKD, MKR)     & AED   & $\mathcal{L}_1$    & (AKD, MKR)    & AED   & $\mathcal{L}_1$    & AKD      & AED   & $\mathcal{L}_1$     \\ \midrule
    X2Face\cite{x2face}     & 0.080 & (17.65, 0.109) & 0.27  & -     & -             & -     & 0.078 & 7.69     & 0.405 & -      \\
    Monkey-Net\cite{monkey} & 0.077 & (10.80, 0.059) & 0.288 & -     & -             & -     & 0.049 & 1.89     & 0.199 & -      \\
    FOMM\cite{fomm}       & 0.055 & (6.62, 0.031)  & 0.164 & 0.033 & (7.07, 0.014) & 0.163 & 0.041 & 1.29     & 0.135 & 0.0225  \\
    MRAA\cite{mraa}       & 0.048 & (5.41, 0.025)  & \textbf{0.149} & \textbf{0.026} & (4.01, 0.012) & \textbf{0.116} & 0.040 &    1.29    & 0.136 & 0.0274 \\
    Ours       & \textbf{0.045} & (\textbf{4.57}, \textbf{0.018})  & 0.151 & 0.027 & (\textbf{3.39}, \textbf{0.007}) & 0.124 & \textbf{0.039} & \textbf{1.22}     & \textbf{0.125} &  \textbf{0.0212}  \\ \bottomrule
  \end{tabular}
  }
  \caption{Video reconstruction: comparison with the state of the art on four different datasets. $K=10$ for all methods. (Lower is better, best result in bold) }
  \label{table:vr}
  \vspace{-2mm}
  \end{table*}

\section{Experiments}

\subsection{Benchmarks}
We trained on multiple types of datasets, including talking faces, human bodies and pixel animation. The pre-processing and training-test splitting strategies for each dataset are the same as in\cite{mraa}. The datasets are as follows:

\begin{itemize}
\setlength{\itemsep}{0pt}
\setlength{\parsep}{0pt}
\setlength{\parskip}{0pt}
  \item \emph{VoxCeleb}\cite{voxceleb}: Videos of different celebrities talking downloaded from youtube cropped to 256*256 resolution according to the bounding boxes of the faces. The length of the videos ranges from 64 to 1024 frames.
  \item \emph{TaiChiHD}\cite{fomm}: Videos of full bodies TaiChi performance downloaded from youtube cropped to 256*256 resolution according to the bounding boxes of the bodies.
  \item \emph{TED-talks}\cite{mraa}: Videos of TED talk downloaded from youtube, which is a new dataset proposed in MRAA. The videos were downscaled to 384*384 resolution according to the upper part of the human bodies. The length of the videos ranges from 64 to 1024 frames.
  \item \emph{MGif}\cite{monkey}: A dataset of \emph{.gif} files of pixel animations about animals running, which was collected using google searches.
\end{itemize}

In previous work, video reconstruction was used to evaluate the quality of motion transfer by taking the first frame $\mathbf{D}_1$ of a video as the source image to reconstruct $\{\mathbf{D}_t\}, t = 1,2,\dots ,T$. We used the same quantitative metrics:

\begin{itemize}
  \setlength{\itemsep}{0pt}
  \setlength{\parsep}{0pt}
  \setlength{\parskip}{0pt}
    \item $\mathcal{L}_1$: Average $\mathcal{L}_1$ distance between the generated and driving image pixel values.
    \item \emph{Average keypoint distance} (AKD): AKD evaluates the poses of the generated images. We use the same pre-trained detectors for bodies\cite{openpose} and faces\cite{facealignment} as MRAA\cite{mraa} to extract keypoints from the generated and driving images. Then calculate the average distance of the corresponding keypoints.
    \item \emph{Missing keypoint rate} (MKR): The proportion of keypoints extracted from the pre-trained model\cite{openpose,facealignment} that are present in the driving image but missing in the generated image.
    \item \emph{Average Euclidean distance} (AED): AED evaluates the identity of the generated images. We use the same pre-trained re-identification networks for bodies\cite{reid_body} and faces\cite{reid_face} as MRAA\cite{mraa} to extract identities from the generated and driving images. Then calculate the average $\mathcal{L}_2$ distance of the extracted identity pairs.
    \vspace{-0.4mm}
  \end{itemize}

\vspace{-1.6mm}
\subsection{Comparison}
\vspace{-1.4mm}
We used two GeForce RTX 3090 GPUs to train 100 epochs on each dataset. VoxCeleb, TaiChiHD, and MGif for three days, while TED-talks was trained for eight days due to higher resolution. We compared our method with the current state-of-the-art unsupervised motion transfer method, MRAA\cite{mraa}, on video reconstruction and image animation tasks. We also compared with other baseline methods FOMM\cite{fomm}, Monkey-Net\cite{monkey} and X2Face\cite{x2face} on video reconstruction.
FOMM\cite{fomm}, MRAA\cite{mraa} and our methods are trained for the same number of epochs with $K$ = 10.

\begin{figure}[!t]

  \centering
    \includegraphics[width=0.9\linewidth]{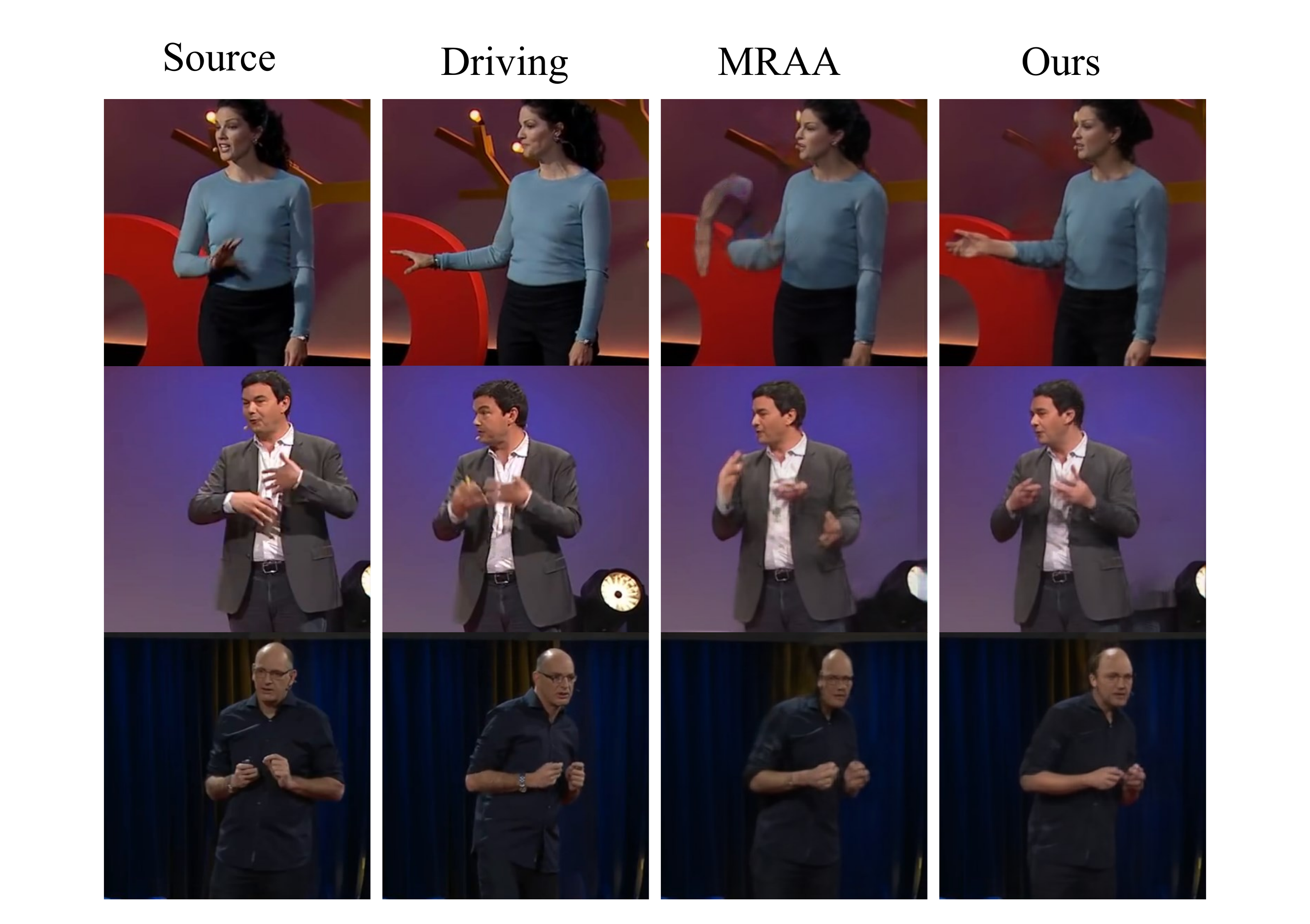}
    \vspace{-2mm}
    \caption{Some bad cases of MRAA\cite{mraa}, while our method shows high quality on video reconstruction task.}
    \label{fig:compare_ted}
    \vspace{-3mm}
\end{figure}

\begin{figure}[!t]
  \centering
    \includegraphics[width=0.9\linewidth]{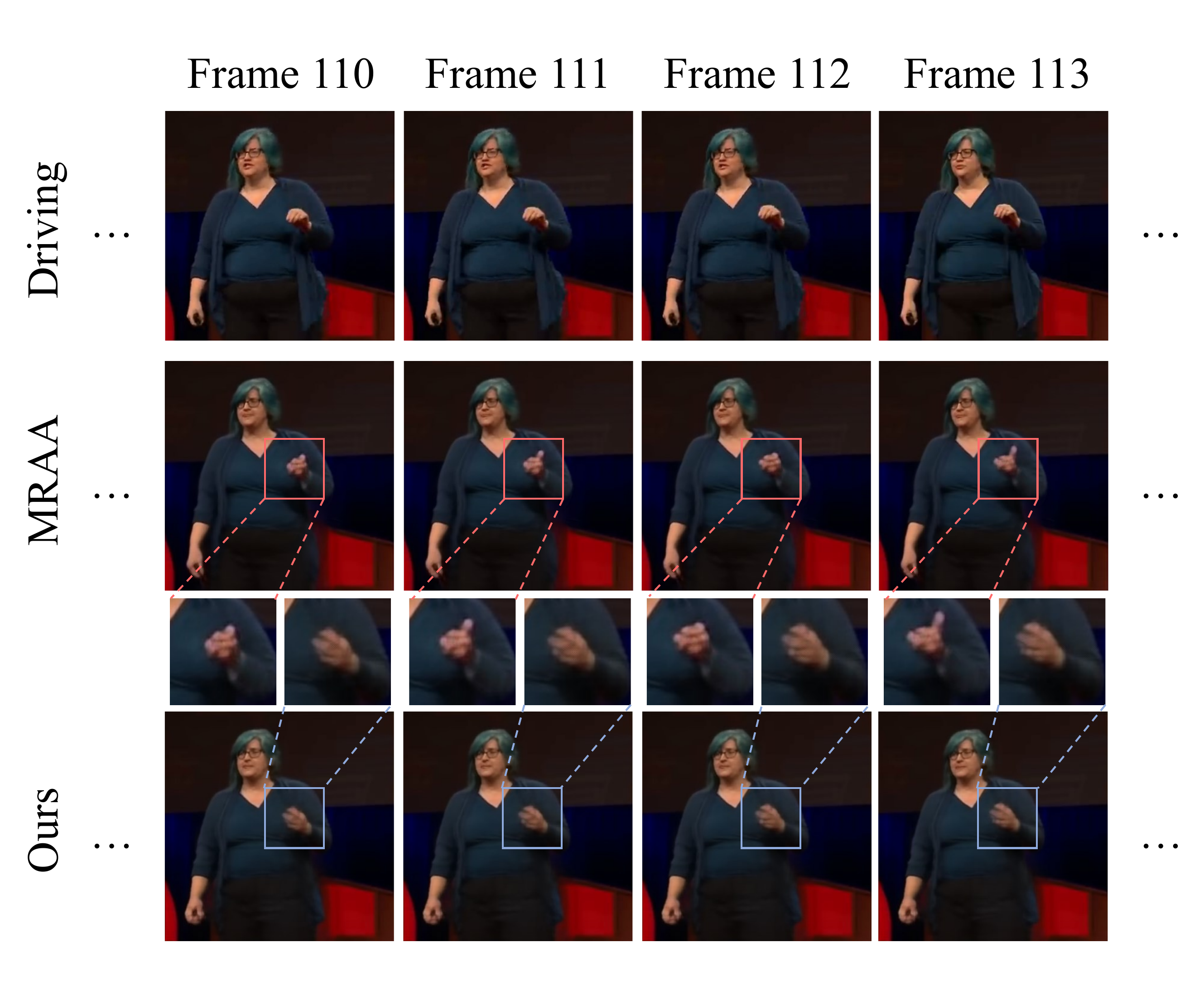}
    \vspace{-2mm}
    \caption{Our method has better temporal continuity than MRAA\cite{mraa} on the video reconstruction task.} 
    \label{fig:lianxu}
    \vspace{-2mm}
\end{figure}

\noindent
\textbf{Video reconstruction}. Quantitative results of video reconstruction are shown in \cref{table:vr}. Our method reaches state-of-the-art results on VoxCeleb, TaiChiHD and MGif datasets, with significant improvements in motion-related metrics on TaiChiHD dataset (15.5\% for AKD and 28.0\% for MKR). This suggests that our method estimates motions more accurately than others. Our method also outperform MRAA\cite{mraa} on TED-talks dataset with motion-related metrics (AKD), but was slightly worse in other metrics (L1 and AED). The latter metrics are related to identity. \cref{fig:compare_ted} shows several bad cases in MRAA\cite{mraa} on TED-talks dataset while our method has better reconstruction quality in hand, arm, and head areas. 

Another advantage of our method is that the reconstructed video has a better temporal continuity. MRAA\cite{mraa} uses local affine transformations near the keypoints to estimate the motion. Therefore, the temporal continuity of the reconstructed video depends on the smoothness of the keypoints change. If the location of the keypoints in two adjacent frames change greatly, it will cause pixel jitter (as shown in \cref{fig:lianxu}, the video reconstructed by MRAA\cite{mraa} has a redundant finger at frames 111 and 113, but not at frames 110 and 112). Instead, we use TPS motion estimation and each transformation is generated by multiple keypoints, which increases the robustness of motion estimation.

\begin{figure*}[t]
  \centering
    \includegraphics[width=0.87\linewidth]{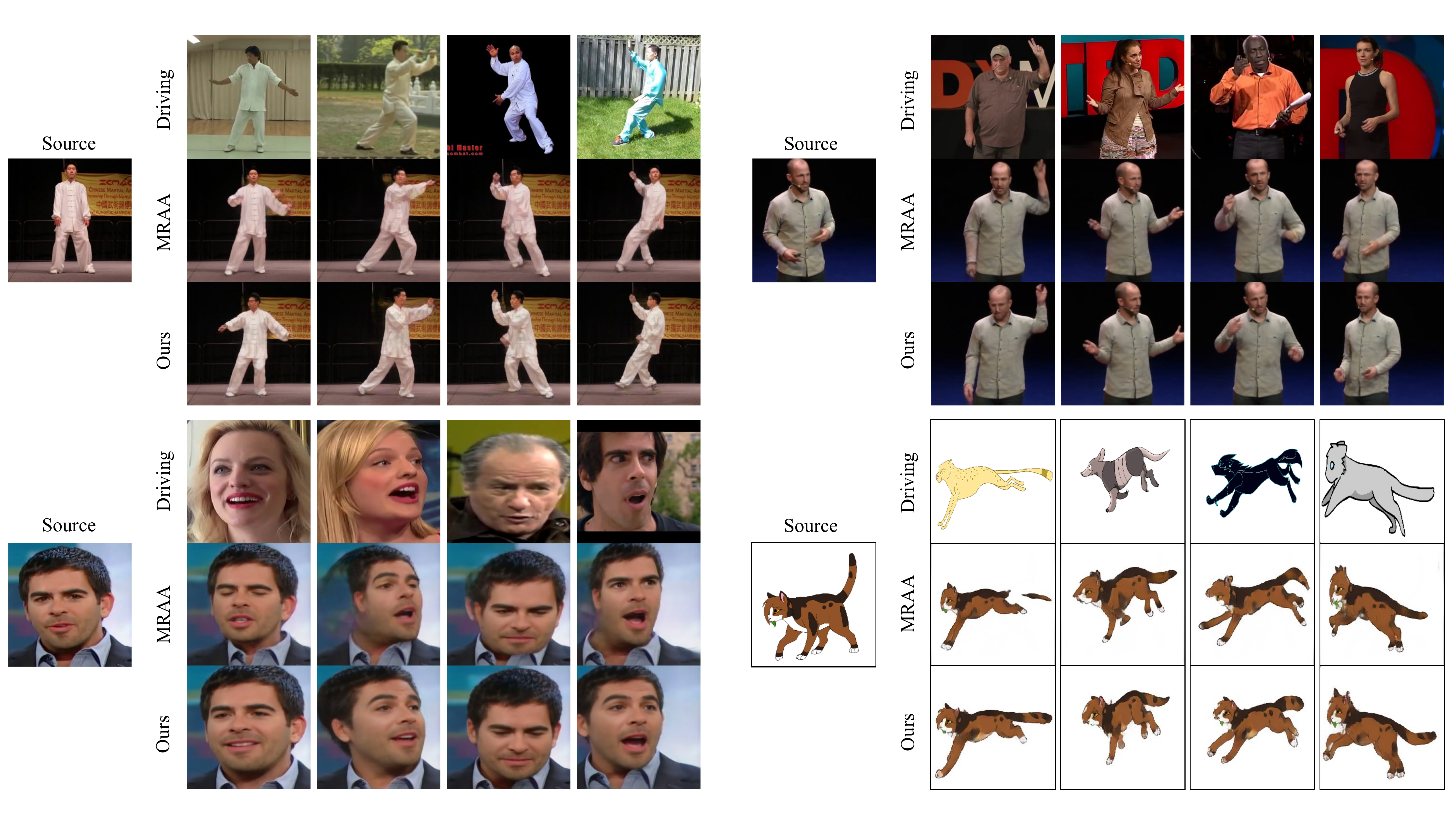}
    \vspace{-2mm}
    \caption{Qualitative comparison with MRAA\cite{mraa} on image animation task: TaiChiHD (top left), TED-talks (top right), VoxCeleb (bottom left), MGif (bottom right).}
    \label{fig:animation}
    \vspace{-3mm}
\end{figure*}

\begin{table}[t]
  \centering
    \setlength{\tabcolsep}{6mm}{
  \begin{tabular}{c|cc}
  \toprule
            & Continuity & Authenticity \\ \midrule
  TaiChiHD  &   71.30\%  &     86.14\%  \\
  TED-talks &   63.93\%  &     58.44\%  \\
  VoxCeleb  &   80.95\%  &     61.54\%  \\
  \bottomrule
\end{tabular}
  }
  \vspace{-1mm}
\caption{User study on image animation, numbers respect the proportion (\%)
of users that prefer our method over MRAA\cite{mraa}.}
\label{table:user}
\vspace{-2.5mm}
\end{table}

\noindent
\textbf{Image Animation}. \cref{fig:animation} shows selected image animation results of our method compared with MRAA\cite{mraa} on the four datasets. Both MRAA\cite{mraa} and our method use the \emph{avd} mode to generate image animation. The results show that our method has better motion transfer performance for human bodies (TaiChiHD and TED-talks), but the ability to maintain image details such as clothes and faces is slightly poor. For human faces (VoxCeleb), ghosts will appear in the images generated by MRAA\cite{mraa}, while our method has better generation quality. Our method also performs better than MRAA\cite{mraa} on pixel animations (MGif). However, when the size of the identity differs greatly, neither MRAA\cite{mraa} nor our method works well (for example, animate a puppy according to the motion of a giraffe).

In order to quantitatively evaluate the quality of image animation, we designed a questionnaire containing randomly selected 20 pairs of videos generated by our method and MRAA\cite{mraa} on TaiChiHD, TED-talks and VoxCeleb for user preference study. People were invited to judge which of the two videos was better for continuity and authenticity. Results are reported in \cref{table:user}. Our method performs much better on temporal continuity than MRAA\cite{mraa}, and most users prefer the videos generated by our method for authenticity.

\begin{table*}[!t]
  \centering
  \setlength{\tabcolsep}{2.5mm}{
     
  \begin{tabular}{c|ccc|ccc|ccc}
     \toprule
       & \multicolumn{3}{c|}{K=5, (K=2 for ours)} & \multicolumn{3}{c|}{K=10, (K=4 for ours)} & \multicolumn{3}{c}{K=20, (K=8 for ours)} \\
       & $\mathcal{L}_{1}$   & (AKD, MKR)  & AED  & $\mathcal{L}_{1}$   & (AKD, MKR)   & AED  & $\mathcal{L}_{1}$   & (AKD, MKR)  & AED  \\ \midrule
  FOMM\cite{fomm} &   0.062   &   (7.34, 0.036)  &  0.181     &   0.056   &   (6.53, 0.033)  &   0.172 &  0.062  &  (8.29, 0.049)  &  0.196
  \\
  MRAA\cite{mraa} & 0.049 &   (6.04, 0.029)  &   \textbf{0.162}   &   0.048   &    (5.41, 0.025)    &   \textbf{0.149}   &  0.046    &   (5.17, 0.026)   &   \textbf{0.141}    \\
  Ours &   \textbf{0.048}   &    (\textbf{5.24}, \textbf{0.022})        &    0.166  &   \textbf{0.046}   &   (\textbf{4.84}, \textbf{0.020})          &  0.156    &   \textbf{0.045}   &     (\textbf{4.67}, \textbf{0.019})       &    0.150 
  \\ \bottomrule
  \end{tabular}
  }
  \vspace{-0.5mm}
  \caption{Additional experiments on TaiChiHD for different K and similar bottleneck sizes for MRAA and ours. (Best result in bold.)}
  \label{table}
  \vspace{-1.5mm}
\end{table*}

\begin{table}[!t]
  \centering
    \setlength{\tabcolsep}{3.1mm}{
  \begin{tabular}{c|ccc}
    
    \toprule
                  & $\mathcal{L}_1$     & (AKD, MKR)    & AED   \\ \midrule
  MRAA\cite{mraa}            & 0.048 & (5.41, 0.025)  & \textbf{0.149}  \\ 
   TPS         & 0.048  & (4.96, 0.020) & 0.153 \\
  $+$ Dropout        & 0.048  & (4.66, 0.018) & 0.156 \\
  $+$ Multi-Masks & 0.046  & (4.73, 0.018) & 0.150 \\
  $+ \mathcal{L}_{bg}$     & 0.046 & (4.64, 0.020) & 0.151 \\
 $+ \mathcal{L}_{warp}$   & \textbf{0.045}  & (\textbf{4.57}, \textbf{0.018})  & 0.151 \\ \bottomrule
  \end{tabular}
  }
  \caption{Ablation study for video reconstruction on TaiChiHD. (Lower is better, best result in bold)}
  \label{table:ablation}
  \vspace{-3mm}
  \end{table}

\vspace{-1.5mm}
\subsection{Ablations}
\vspace{-1.5mm}
We perform ablation experiments on TaiChiHD dataset to demonstrate the improvement from each of our proposed components. We add our components in turn and compare the video reconstruction metrics with MRAA\cite{mraa}.  Firstly, we use TPS transformations to estimate the optical flow instead of local affine transformations in MRAA\cite{mraa}. Then we added the dropout operation and the multi-resolution occlusion masks. Finally, we add $\mathcal{L}_{bg}$ and $\mathcal{L}_{warp}$ during training. Quantitative results are shown in \cref{table:ablation}.

\begin{figure}[!t]
  \vspace{-2mm}
  \centering
    \includegraphics[width=1\linewidth]{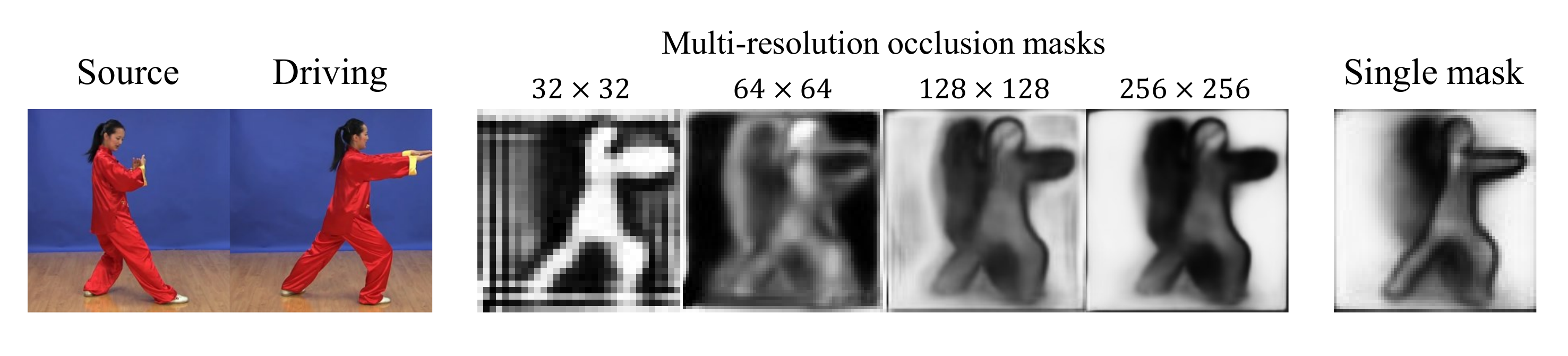}
    \caption{Comparison of learned multi-resolution occlusion mask and single occlusion mask.}
    \vspace{-2.5mm}
    \label{fig:masks}
    
\end{figure}

The second row of \cref{table:ablation} demonstrates that TPS motion estimation improves AKD and MKR, resulting in more accurate motion estimation. Comparing the second and third rows of \cref{table:ablation}, dropout can bring lower AKD and MKR, which indicates that dropout makes each TPS transformation contributes to the optical flow to distort the object in $\mathbf{S}$ into a more accurate pose. However, dropout also affects AED because the more complex optical flow means the larger area of missing regions in warped feature maps, resulting in insufficient information for the Inpainting Network to revise the image. The fourth row of \cref{table:ablation} shows that the multi-resolution occlusion masks bring an improvement to $\mathcal{L}_1$ and AED, which can help the Inpainting Network to generate higher quality images. \cref{fig:masks} shows the multi-resolution occlusion masks we learned by our full method, compared with the single occlusion mask learned by the method in the third row of \cref{table:ablation}. But at the same time, it brings a higher AKD, which is not what we expected. When we add $\mathcal{L}_{bg}$ and $\mathcal{L}_{warp}$, AKD gradually decreases, the full method can achieve a better balance on the four metrics.

As with MRAA and FOMM, one of the most important hyper-parameters in our model is the number of TPS transformations, $K$, which corresponds to the dimension of the motion representation.
The dimensions of FOMM and MRAA are $K*(2+4)$ and $(K+1)*(2+4)$, while ours is $K*(6+5*2)+6$. When K = 5, 10, and 20, the dimensions of MRAA are 36, 66, and 126, respectively. We set K of our method to be 2, 4, and 8 to compare with them, and the dimensions are 38, 70, and 118, respectively, which is similar to MRAA.
\cref{table} shows the experiment results, which demonstrates that our method can achieve better motion-related metrics than MRAA when using similar motion description dimensions.

\vspace{-1mm}
\section{Discussion and Conclusion}
\vspace{-1mm}

In this paper, we first discuss the drawbacks of using local affine transformations to approximate motions in previous works and propose TPS motion estimation to estimate an optical flow that warps the feature maps of the source image to the feature domain of the driving image. In addition, we use dropout for TPS transformations before combining them in the early stage of training, which keeps the network from excessive reliance on a few TPS transformations and avoids the network falling into local optimums. Secondly, the multi-resolution occlusion masks are used to achieve more effective feature fusion instead of a single occlusion mask. Finally, we design additional auxiliary loss functions and proved experimentally effective. 

Our method achieves state-of-the-art performance on most benchmarks with visible improvements in motion-related metrics. However, our approach does not perform well when an extreme identity mismatch occurs. Unsupervised motion transfer remains a worthwhile challenge.

\noindent
\textbf{Potential negative societal impact}. 
While the proposed method may be used to make fake videos for spoofing, some detection software will easily determine the authenticity of videos by analyzing the color textures\cite{anti} or using the depth information obtained through the proximity sensors, which cannot be simulated by 2D image generating methods. And our approach can create a new benchmark for face anti-spoofing researches\cite{anti_survey}.

{\small
\bibliographystyle{ieee_fullname}
\bibliography{egbib}
}

\end{document}